\documentclass[sn-mathphys-num]{sn-jnl}

\usepackage{graphicx}%
\usepackage{multirow}%
\usepackage{amsmath,amssymb,amsfonts}%
\usepackage{amsthm}%
\usepackage{mathrsfs}%
\usepackage[title]{appendix}%
\usepackage{xcolor}%
\usepackage{textcomp}%
\usepackage{manyfoot}%
\usepackage{booktabs}%
\usepackage{algorithm}%
\usepackage{algorithmicx}%
\usepackage{algpseudocode}%
\usepackage{listings}%

\theoremstyle{thmstyleone}%
\theoremstyle{thmstyletwo}%

\theoremstyle{thmstylethree}%

\begin{document}

\title[Article Title]{Explainable Artificial Intelligence and Multicollinearity : A Mini Review of Current Approaches}


\author*[1,2,3,4]{\fnm{Ahmed M} \sur{Salih}}\email{a.salih@leicester.ac.uk}

\affil*[1]{\orgdiv{Department of Population Health Sciences, University of Leicester, University Rd, LE1 7RH, Leicester, UK}}
\affil[2]{\orgdiv{William Harvey Research Institute, NIHR Barts Biomedical Research Centre, Queen Mary University of London, Charterhouse Square, London, EC1M 6BQ, London, UK}}
\affil[3]{\orgdiv{Barts Heart Centre, St Bartholomew’s Hospital, Barts Health NHS Trust, West Smithfield, London, EC1A 7BE, UK}}
\affil[4]{\orgdiv{Department of Computer Science, University of Zakho, Duhok road, Zakho, Kurdistan, Iraq}}


\abstract{Explainable Artificial Intelligence (XAI) methods help to understand the internal mechanism of machine learning models and how they reach a specific decision or made a specific action. The list of informative features is one of the most common output of XAI methods.  Multicollinearity is one of the big issue that should be considered when XAI generates the explanation in terms of the most informative features in an AI system. No review has been dedicated to investigate the current approaches to handle such significant issue. In this paper, we provide a review of the current state-of-the-art approaches in relation to the XAI in the context of recent advances in dealing with the multicollinearity issue. To do so, we searched in three repositories that are: Web of Science, Scopus and IEEE Xplore to find pertinent published papers. After excluding irrelevant papers, seven papers were considered in the review. In addition, we discuss the current XAI methods and their limitations in dealing with the multicollinearity and suggest future directions..}




\maketitle

\section{Introduction}\label{sec1}

Explainable Artificial Intelligence (XAI) emerged to uncover the vagueness around complex models including deep learning and make them more understandable from human point of view~\cite{angelov2021explainable}. It is a set of tools, framework and algorithms that is developed to interpret the predictions of the machine learning models. It has shown massive success in different domains including healthcare~\cite{salih2023explainable}, genetic~\cite{novakovsky2023obtaining}, education~\cite{chaushi2023explainable}, finance~\cite{weber2023applications}, social science~\cite{johs2022explainable} and ecology~\cite{ryo2021explainable}.\\
One of the most common outcome of XAI methods is the list of informative features in the model when applied to tabular data. It shows what are the features that significantly drive the model decision. In addition, it shows whether the association of these features with the outcome positive, negative or does not have effect at all. Other form of XAI outcome is the dependency between a feature or group of features with the outcome. It shows whether the association is linear or non-linear including monotonic.\\
However, XAI is not mature yet and still in the development stage which necessitate to evaluate its outcomes. There are several concerns related to understanding  and interpreting the outcome of XAI methods include the impact of one feature in the model on the outcome when the features are not independent. One of the big issue in machine learning and XAI is presence of multicollinearity. It is a phenomena where multiple features in the machine learning models are highly correlated. Such issue affect the reliability of the prediction and interpreting the results. This is because it is difficult to determine the individual effect of a feature in the model on the outcome. In real-life scenario and especially in healthcare and biology, the features are usually correlated. XAI methods need to consider this phenomena when providing list of most informative features and when they reveal the impact of one individual feature toward the outcome.\\
This review aims to explore the advances and solutions related to issue of multicollinearity in XAI methods. Moreover, it discussed some common XAI methods and how they deal with the multicollinearity issue when they explain a model.
\section{XAI and Multicollinearity}
On of the most common output of XAI methods is the list of informative features which sort them in a descend order based on their influence on the model prediction. For instance, if feature\textit{X} is on the top of the list, it means that feature has the most powerful impact on the prediction. This is correct if the feature is independent from other features in the model. However, if feature \textit{X} is highly collinear with other features in the model, the interpretation of the list of informative features is not realistic and not reliable. Other form of XAI method is the dependency plot where it shows the association of one or two features toward model output. It shows how change in a feature led to change in the output. Again, this seems incorrect if that feature is collinear with other features in the model. Below we discuss most common XAI methods that provide list of informative features or reveal the kind of association between a feature and the outcome. We do not discuss in deep the technical details of each method, we rather discuss how each method deals with the multicollinearity issue when it explains the model prediction.
\begin{enumerate}
    \item \textbf{SHapley Additive exPlanations (SHAP)} is a mode-agnostic XAI based on game theory to calculate the attribution of each feature in the model toward model prediction~\cite{NIPS2017}. There have been many publications that explained in details how SHAP works which is outside the scope of the current review. However, what we are interested in is the assumption behind features-dependency. In the original version of SHAP, it assumes that the features are independent when it calculates a score of each feature. For instance, if feature \textit{X} and \textit{Z} are collinear and they are correlated with the outcome, only one of them get a high SHAP score. This is because the second one does not improve the model performance when it is added in the model. Accordingly, the SHAP score does not really capture the real impact of the features toward the outcome when the features are collinear. There have been many methods proposed to modify SHAP in order to provide explanation and consider the multicollinearity simultaneously~\cite{aas2021explaining, mase2019explaining, basu2022multicollinearity, redelmeier2020explaining}. The proposed methods will be discussed more in details in the following section.
    \item \textbf{Local Interpretable Model-agnostic Explanations (LIME)} is a local model-agnostic XAI method. It approximates any model to explain an instance through a local linear model. The outcome of LIME is the coefficient (weights) value of the features generated by the local linear model. The interpretation of the weights indicates that increasing or decreasing one unite in the feature causes increasing or deceasing in the outcome by one unite while holding all other features in the model constant. Such interpretation is correct when the features are independent. However, in real life applications the features are usually collinear. Accordingly, LIME explanation is not realistic as it does not consider multicollinearity in its outcomes.
    \item \textbf{Partial Dependence Plot (PDP)} is a model-agnostic XAI method that reveals the impact of one or two features on the outcome in machine learning models~\cite{friedman2001greedy}. It shows whether the association between the features simple (e.g., linear), monotonic or more complicated. One of the limitations of PDP is that it assumes the features are independent when it generates the plots~\cite{molnar2018guide}. It is kind of similar to interpreting the coefficient in a linear model because it averages over the marginal distribution of other features when it generates the plot for a specific feature. The explanation provided by PDP does not reflect the interactions between the features and on the outcome.
    \item \textbf{Accumulated Local Effects (ALE)}~\cite{apley2020visualizing} is similar to PDP which aims to examine the association between one or two features with the outcome globally. However, ALE is better in terms of handling the multicollinearity issue because it focuses on the prediction change rather than on averaging the features as the case in PDP. It divides the feature of interest into intervals and then calculate the changes in the prediction in each interval. The plots generated by ALE are more realistic and unbiased compared to PDP. However, the interpretation of the plots assume that the changes happen in the prediction for each interval while the instances value of other features are fixed. Moreover, the presence of  multicollinearity make it difficult to understand the ALE plot and interpret it correctly~\cite{molnar2018guide}.
    \item \textbf{Anchor}~\cite{ribeiro2018anchors} is a local agnostic rule-based XAI method that aims to explain a classification model for an individual. If the changes does not affect the prediction, a rule anchors the prediction. Thereafter, these rules are presented in if-then statement which is easy to understand. The method has some limitations in terms of highly configurable, discretization  and running time. To decrease the running time, anchor uses reinforcement learning alongside with graph search analysis. The anchors might be difficult to be build accurately when the features are dependent because it is perturbation-based. 
\end{enumerate}
\section{Literature review}
\begin{figure}[ht!]
\centering
\includegraphics[height=8 cm]{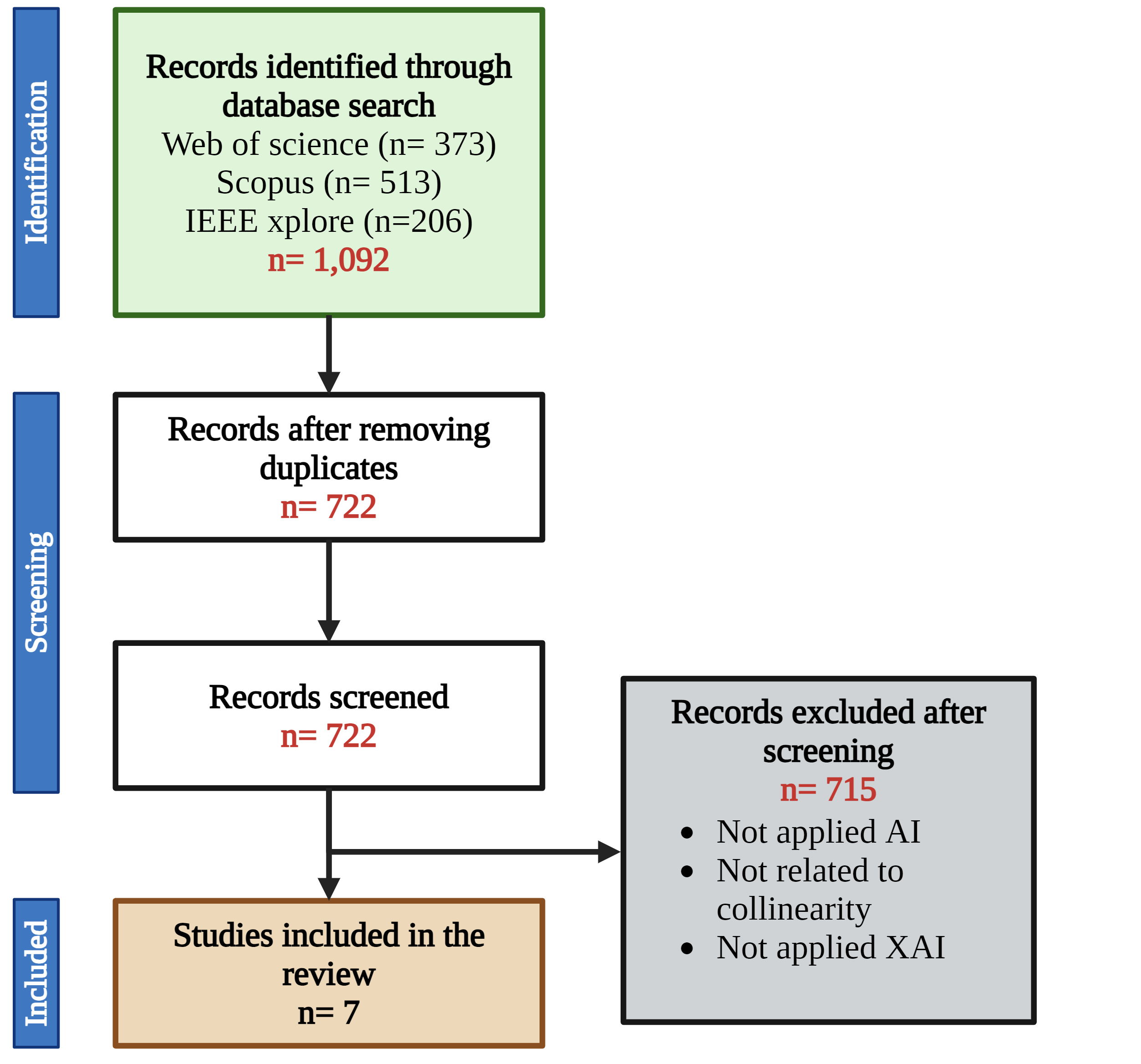}
\caption{Literature review search.}\label{LRS}
\end{figure}

Search was conducted on 21\textsuperscript{st} of April 2024 in the Web of Science, Scopus and IEEE Xplore repositories. The search was specified to English papers without restriction in the year of publication. The search was conducted using a query that involves the key words and synonyms related to multicollinearity and XAI methods (Table S1). The search aimed to find the key words in the abstract of the papers.\\
Figure~\ref{LRS} shows the number of papers from each repositories and the process of excluding papers. Any paper that did not use XAI dealing with multicollinearity issue was excluded. There was a massive number of papers when the search was conducted. However, after removing the duplicates, excluding those did not deal with XAI and  multicollinearity, seven papers left to include in the current review.
Table~\ref{compare} lists the seven papers included in the review. The table shows whether the method can be implemented as a global or local explanation. In addition, it shows whether the proposed method specific to an XAI method or agnostic which means can be applied to any XAI method. \\
\begin{table}[ht!]
\centering
\begin{tabular}{|l|c|c|c|}
\hline
\textbf{Method}  & \textbf{Global or Local} &\textbf{Specific or Agnostic} \\ \hline
Modified Index Position (MIP)~\cite{salih2024characterizing}            &Global \& Local                 & Agnostic\\ \hline
Extended Kernel SHAP~\cite{aas2021explaining}                           & Local                          &  Specific to SHAP\\ \hline
Normalized Movement Rate (NMR)~\cite{salih2022investigating}            &Global \& Local                 & Agnostic \\ \hline
SHAP Cohort Refinement (SCR)~\cite{mase2019explaining}                  & Local                          &  Specific to SHAP\\ \hline
Multi$-$collinearity Corrected (MCC)~\cite{basu2022multicollinearity}   & Global                         &  Specific to SHAP\\ \hline
Conditional Subgroups (CS)~\cite{molnar2023model}                       & Global                         & Agnostic \\ \hline
Conditional Inference Trees (CIT)~\cite{redelmeier2020explaining}       &Local                           &  Specific to SHAP \\ \hline
\end{tabular}
\caption{The proposed methods to handle the features-dependency issue in the outcome of XAI.}
\label{compare}
\end{table}
The seven methods are discussed more in details below. The discussion is focused on how it works, applied to any XAI method and whether its implementation is publicly available in any coding language so the reader could use it.
\begin{enumerate}

\item \textbf{Modified Index Position}: MIP~\cite{salih2024characterizing} was proposed to modify and enhance any XAI method that does not consider multicollinearity when generates the explanation. It can be applied to any XAI method both globally and locally. The method works as post-hoc of XAI to consider features-dependency issue. Once the model was trained and XAI was applied to generate the list of significant features, then MIP modifies the outcomes of XAI to consider the multicollinearity. It removes iteratively the top feature from the list, retrain the model, apply XAI and generates the list of significant features again. The hypothesis behind that is if two features are collinear and correlated with the outcome, then once one of them removed from the model, the other one will top the list. The faster the feature reach the top list when every time one is removed, the more the feature is significant in the model. They evaluated their method using Principal component analysis applied to the features to generate new independent components. Their method shows that when PCA was applied, the outcome of XAI more stable and less model-dependent. The implementation of the MIP in Python is available at \href{https://github.com/amaa11/MIP}{Github}.
\item \textbf{Extended Kernel SHAP}: The method was mainly proposed to modify the Kernel SHAP method to consider the multicollinearity and explain the model locally at individual level~\cite{aas2021explaining}. In the original Kernal SHAP method, it is assumed that the features are independent and the conditional distribution can be replaced by the marginal distribution. However, such assumption is not realistic in real-life applications specifically in healthy-care which might lead to incorrect explanation. Extended Kernel SHAP proposed four approaches to estimate the conditional distribution instead of replacing it with the marginal distribution. The proposed approaches are Gaussian copula distribution, Gaussian distribution, Empirical conditional distribution and the combination of the empirical approach and either the Gaussian or the Gaussian copula. The user has the ability to chose a specific distribution over others based on the distribution of the data. The method is a local explanation and works only with SHAP. The implementation of the method in R is available at \href{https://norskregnesentral.github.io/shapr/}{Github}.
\item \textbf{Normalized Movement Rate} : XAI methods are usually model-dependent which means each model might provide different explanation even if the accuracy is relatively similar~\cite{salih2023commentary}. One of the main factor that causes this discrepancy is the multicollinearity because each model deals with this issue differently. The question is when applying several models which one of the explanation to consider given that the accuracy of the models is similar. NMR~\cite{salih2022investigating} was proposed to assess the stability of the explanation in presence of multicollinearity. It works by iteratively removing the top feature from the outcome of XAI, then train the model again and apply XAI. Thereafter, it compares whether the left features changed their positions in the XAI outcome compared to previous list. The less the features change their positions, the better the model against the collinearity. It calculates a value between zero and one which represents the changes of the positions of the features when every time the top one is removed. NMR value close to zero means that the features do not change their positions a lot when the top one is removed from the model and the XAI is more stable. On contrary, NMR value close to one means that when the top feature is removed from the mode, the rest of the features change their positions massively which indicates that the model and the XAI outcome is not stable or robust against the multicollinearity. NMR can be applied to any XAI both globally and locally. It does not generate an explanation with considering multicollinearity issue, it rather help to chose a more stable model against features-dependency issue. The implementation of the NMR in Python is available at \href{https://github.com/amaa11/NMR}{Github}.
\item \textbf{SHAP Cohort Refinement}: SCR~\cite{mase2019explaining} is another method was proposed to deal with the limitation of the original version of SHAP and consider the multicollinearity in the explanation locally for a single subject. In order to deal with the issue, it creates a new cohort that is similar to the instance need to be explained. To do so, it combines many data points that are similar to the instance that need to be explained. Measuring the similarity depends on the kind of the data. If the data are categorical, then SCR uses the frequency of each value to find similar instances. On the other hand, if the features are continues variables, then it uses distance-based methods. Once it creates the cohort, then it applies SHAP to explain the target instance. Accordingly, the method eliminates or mitigates the impact of the multicollinearity as the explanation of the instance was explained in reference to the new similar cohort. The implementation of the SCR in Python is available at \href{https://github.com/cohortshapley/cohortshapley}{Github}.
\item \textbf{Multi$-$Collinearity Corrected}: Another approach was proposed to handle the multicollinearity issue in the explanation provided by SHAP~\cite{basu2022multicollinearity}. The main idea behind MCC is that they remove the impact of dependency between the features before calculating the SHAP score. To do so, they proposed to add an Adjustment Factor (AF) for each feature correlated with the feature of interest while removing or randomize the feature that SHAP score is calculated for. The role of the AF is to decrease or eliminate the correlation between the feature of interest with the rest of features. This is validated by calculating the covariance between the feature of interest with the other features plus the AF. MCC was implemented in the approximate SHAP version as the original SHAP is computationally expensive. They implemented the method on several datasets using several models.
\item \textbf{Conditional Subgroups}: CS~\cite{molnar2023model} was proposed to handle the issue of multicollinearity using conditional subgroup with permutation feature importance and partial dependency plot. The main idea behind the proposed method is that creating groups in a way that the feature of interest is less dependent on all other features in the all subgroups. Decision tree is used to split the data into groups in a way that the distribution of the feature of interest more homogeneous in one group and more heterogeneous in the rest of groups. It is implemented with PDP which lead to a different interpretation of the original version of PDP. In the modified version of PDP, the change in the outcome is reflected by the change of the feature of the interest given that the variation in all other features based on the join distribution. The implementation of the SC in R is available at \href{https://github.com/christophM/paper_conditional_subgroups}{Github}.
\item \textbf{Conditional Inference Trees}: Extended Kernel SHAP~\cite{aas2021explaining} was proposed to extend the original version of SHAP to consider the dependency among the features for continues variables by estimating the dependency distribution. However, the method was limited to only continues features while in real-life scenario a model might involve more than one type of features. CIT~\cite{redelmeier2020explaining} was proposed to extend the SHAP value method aiming to solve the dependency among the features when the model involves mixture of different types of features including continuous, discrete, ordinal and categorical. The authors followed similar approach to~\cite{aas2021explaining} to handle the multicollinearity issue by estimating the conditional distribution. However, instead of using the marginal distribution, they used CIT to estimate the conditional distribution. CIT is known to be better to handle categorical variables and modelling simple and complex collinearity structure. The proposed method is available in R as additional method to the shapr package~\href{https://norskregnesentral.github.io/shapr/}{Github}.
\end{enumerate}
\section{Open issue and future direction}
Our review shows that although XAI methods are applied extensively in variety of research area including sensitive domain where misinterpreting is sever, multicollinearity issue is not explored and investigated adequately in the literature. There is no XAI method that by its nature mitigate the impact of such important issue. The current available XAI methods and their modified versions are either limited to a specific XAI method (e.g., SHAP) or they are limited to local explanation. Moreover, the methods based on the conditional distribution depends highly on the efficiency of the conditional sampler which is challenging in high dimension~\cite{molnar2020general}. Accordingly, more research is required on sampling methods with XAI and presence of multicollinearity. Conditional probability could be one of the ways to estimate the impact of a specific feature given that the other features have specific values. In addition, it might be difficult to estimate the impact of a feature on the outcome by one single score due to the interaction between the dependent-features. Possible way is to estimate a distribution using Bayesian models which reflect the impact of a feature given that different values for different correlated features. Moreover, current plots of XAI to explain the models either locally or globally are misleading because they do not reflect the complexity interaction between the features and they rather give impression of that the features are independent. Accordingly, new plots are required to embed and reflect the interaction between the features and their effect on the outcome. This is especially significant when the models involves more than one type of features and they are highly correlated.

\bibliography{sn-bibliography}

\end{document}